\documentclass{elsarticle}

\usepackage{lineno,hyperref,amsmath}
\usepackage{tikz}
\usepackage{pgfplots}
\usetikzlibrary{calc,trees,positioning,arrows,chains,shapes.geometric,%
    decorations.pathreplacing,decorations.pathmorphing,shapes,%
    matrix,shapes.symbols}

\journal{Journal of Computer Vision and Image Understanding}
\newcommand{\norm}[1]{\left\lVert#1\right\rVert}
\newcommand{\card}[1]{\vert#1\vert}
\newcommand{\argmax}{\operatornamewithlimits{argmax}}

\begin{document}

\begin{frontmatter}
\title{Saliency Driven Object recognition in egocentric videos
with deep CNN\\ \footnotesize Electronic copy submitted to the Journal of Computer Vision and Image Understanding}
\author[bx,labri]{Philippe P\'erez de San Roman\corref{cor1}}\ead{philippe.perezdesanroman@labri.fr}
\author[bx,labri]{Jenny Benois-Pineau\corref{cor2}}\ead{benois-p@labri.fr}
\author[bx,labri]{Jean-Philippe Domenger\corref{cor3}}\ead{domenger@labri.fr}
\author[bx,incia]{Florent Paclet\corref{cor5}}\ead{florent.paclet@u-bordeaux.fr}
\author[bx,incia]{Daniel Cataert\corref{cor6}}\ead{daniel.cattaert@u-bordeaux.fr}
\author[bx,incia]{Aymar de Rugy\corref{cor4}}\ead{aymar.derugy@u-bordeaux.fr}
\address[bx]{Université de Bordeaux, 35 Place Pey Berland, 33000 Bordeaux, France}
\address[labri]{LaBRI - UMR 5800 - CNRS Université de Bordeaux, IPB, 351, cours de la Libération, 33405 Talence, France}
\address[incia]{INCIA - UMR 5287 - CNRS Université de Bordeaux, 146, rue Léo Saignat
33076 Bordeaux cedex, France}
\cortext[cor2]{Corresponding author}

\begin{abstract}
The problem of object recognition in natural scenes has been recently successfully addressed with Deep Convolutional Neuronal Networks giving a significant break-through in recognition scores. The computational efficiency  of Deep CNNs as a function of their depth, allows for their use in real-time applications. One of the key issues here is to reduce the number of windows selected from images to be submitted to a Deep CNN. This is usually solved by preliminary segmentation and selection of specific windows, having outstanding "objectiveness" or other value of indicators of possible location of objects. In this paper we propose a Deep CNN approach and the general framework for recognition of objects in a real-time scenario and in an egocentric perspective. Here the window of interest is built on the basis of visual attention map computed over gaze fixations measured by a glass-worn eye-tracker. The application of this set-up is an interactive user-friendly environment for upper-limb amputees. Vision has to help the subject to control his worn neuro-prosthesis in case of a small amount of remaining muscles when the EMG control becomes unefficient. The recognition results on a specifically recorded corpus of 151 videos with simple geometrical objects show the mAP of 64,6\% and the computational time at the generalization lower than a time of a visual fixation on the object-of-interest. 
\end{abstract}


\end{frontmatter}

\section{Introduction and motivation}\label{Sec:1}

The problem of natural object recognition in images has been in the center of computer vision community since quite a lot of time. Previous PascalVOC challenge \citep{DataSet:PascalVOC} and ongoning ImageNet Large Scale Visual Recognition Challenge (ILSVRC) \cite{DataSet:ImageNet} have united important task forces for finding the solution of this problem in natural visual scenes. Since recently, developed approaches find their place in a  quite realistic application for helath care and patient's monitoring such as in \cite{7423809}. The pioneering works on real-world manipulated object recognition in egocentric perspective \cite{Pirsiavash2012} for evaluation of cognitive impairment of Alzheimer patients \cite{CNN:Saliency} showed that even if the state-of-the art in egocentric object recognition does not allow for getting scores approaching  100\% for all object categories and requires a heavy annotation process, the information on the object of interest the human interacts with is essential for assistance and evaluation of patients and impaired subjects. In this paper we develop an object recognition approach for assistance of upper-limb amputees wearing neuro-prostheses. 

 Classic myoelectric control of neuro-prostheses uses the activity of the remaining muscles to control the multiple degrees of freedom of the prosthesis. This strategy, however, faces a fundamental problem related to fact that the higher the amputation, the higher the number of degrees of freedom of the prosthesis to control with less control signals from the fewer remaining muscles. In addition, all commercially available myoelectric prosthesis are only controlled with two muscle groups, one flexor and one extensor, and therefore involve sequential control of individual joint with unnatural control schemes to switch between joints. The tedious learning for relatively mediocre results associated with these control schemes has motivated several laboratories at developing control schemes that integrate a higher number of muscles, either through pattern recognition of movement classes from muscle recordings \cite{Parker2006,Smith2011}, or through regressions that use muscle activities for simultaneous proportional control of the multiple degrees of freedom of the prosthesis \cite{Farina2014, Hahne2014, Jiang2009}. Despite their merits, however, these attempts do not resolve the key limitation in terms of number of remaining muscles, although some control signals could be recovered using highly invasive surgical techniques such as nerve recording \cite{WodlingerDurand2009} or targeted muscle re innervation \cite{Kuiken2009}. 

Promising and much less invasive alternatives propose to use other control signals such as from computer vision \cite{Markovic2014, Markovic2015}, gaze information \cite{Corbett2012,Corbett2014}, and/or residual biological motion \cite{Kaliki2013}. In \cite{Markovic2014}, stereo-vision from camera integrated in augmented reality glasses was used to automatically select grasp type and size from a visible object, and \cite{Markovic2015} added inertial sensing to automatically align wrist orientation to that of the object to grasp. In both instances, however, only one object was present in front of the subject, thereby avoiding the critical problem of recognizing the object of interest from the multiple objects typically present in a natural environment. Furthermore, although camera on glasses provided egocentric videos, gaze information was not available and therefore not used to assist this process. In \cite{Corbett2012,Corbett2014}, gaze information was used, but to supplement muscle recordings for the control of reaching actions rather than to recognize the object of interest. Furthermore, the setup was such that eye tracking was working on a fixed head, and the reaching was on a two dimensional screen. Our goal here is to combine gaze information with recent progresses in deep learning for computer vision, in order to improve real time object recognition from egocentric videos, the long term goal being to incorporate this information into prosthetics control. The rest of the paper is organized as follows. In Section \ref{Sec:2} we analyse the related works in natural object recognition and localization with Deep Convolutional Neural Networks and summarize our contributions, in Section \ref{Sec:3} we present our general approach for saliency driven object recognition with Deep CNN, which uses gaze information. In Section  \ref{Sec:4} we focus on Network design and tunning. Section \ref{Sec:5} presents experiments and results. In Section \ref{Sec:5} discussion, conclusion and further perspectives of this work are presented. 

\section{Related Works}\label{Sec:2}

The problem we adress consists in both: i) object recognition and ii) object localization. In \citep{CNN:RBCNAODS} a good analysis of recent  approaches for object localization has been proposed, such as "regression approaches" as in \cite{CNN:SermanetEZMFL13}\citep{CNN:AgrawalGM14},  and "sliding window approaches" as in \cite{CNN:SermanetKCL13} when the CNN processes multiple overlapping windows. The authors of \cite{CNN:RBCNAODS} propose a so called  Region-based convolutional network (R-CNN). Inspired by "selective search" approach \cite{Loc:SelectiveSearch} it evaluates multiple (2K) "object proposals" and finally train an SVM for classification. 
Such "proposals" are numerous and the multi-task classification with CNNs requires heavy computations. Hence in \cite{CNN:RBCNAODS} the authors report that R-CNN can take from 10s to 45s for object classification. 
Several attempts has been made to accelerate the generalization process. In \cite{CNN:OquabBLS15} a weakly supervised training scheme  is designed. They train  the network on the whole labeled image. The deepest layers of the CNN supply features. Then  fully connected layers - adaptation layers  are proposed considered as convolution layers. Max pooling supplies the scores for different positions of objects. 
In order to reduce the number of object proposals, the spatial grouping of windows was proposed in SPPnet\citep{CNN:HeZR015}. Here the spatial pyramid pooling layer is added above the convolutional layers, transforming the output into fixed-size vectors. Thus built the network not only copes with different sizes of input images, but is from 24 to 54 times faster than AlexNet \citep{CNN:ImageNet} with only 530 ms per image.  
Further acceleration at testing step is proposed in  fast R-CNN \cite{CNN:Girshick15}. Here  the idea of training on the whole image is re-used. The network first processes the whole
image at the deep layers of the network and then, at the uppaer layers, object proposals are processed. Indeed the whole image is used through several convolution and max pooling layers instead of object proposals.  The latter are used at the so-called region-of-interest (ROI) pooling layer. Here a fixed-lentgh feature vector is extracted for each object proposal from the obtained full-image feature map. The feature vectors are then submitted into a sequence of fully connected layers, finally two output layers produce the softmax probability for object classes and the background and the estimate of window corner positions with regression. The speed-up of computation at training step is achieved due to the late selection of features for object proposals, while at the test step, they use truncated SVD decomposition on fully connected layers to accelerate computations for them. The computational time of 600ms is reported per image at the test step.  
All these methods developed for object recognition and localization with Deep CNN aim at increasing mAP and reducing computational time. They work in an "unconstrained" setting, which means that there is no initial assumption on object location. Thus the localization process has to be accelerated. 
One of the trends strongly present in the research in object recognition which also inspired  "selective search" \cite{Loc:SelectiveSearch} approach, consists in using visual saliency of regions, which serves to select object proposal candidates. Out of the methodological framework of Deep CNN, such methods were proposed for the popular Bag-of-Visual-Words (BoVW) model\citep{BoVW:ORLVFM}. The latter served  as object signature but was built on the whole image from qunatized features weighted by underlying saliency value\cite{Sal:Carvalho2012}, \citep{Sal:Gonzalez-DiazBB16}. Despite we have developed a complete saliency-based methodology for all steps in feature engineering approach, such as feature selection, encoding, pooling, and despite its good performances surpassing the most popular state-of-the-art model DPM \citep{DPM}, the BoVW approach showed its limits even with ideal saliency maps such as manually annotated bounding boxes \cite{Sal:STCSM}. As the predicted (objective) or ideal (subjective), computed with gaze-fixations from eye-tracker recording, maps confine image analysis process and eliminate clutter, it is natural to try to incorporate them into the winner model today, such as Deep CNN.
 
The contributions of our work are the following: i) we introduce ideal saliency maps, recorded with head-mounted eye-tracker into "object proposal" selection for our object recognition need in an interactive environment. Taking into account that object recognition has to be conducted, on acting person, we use established facts from cognitive sciences and bio-physicscs to select sequences from video and filter-out distractors ii) we re-use "ImageNet" architecture, propose  addequate data sampling and augmentation relevant to our egocentric setup and show that without any supplementary efforts, the base-line ImageNet allows to get rather good scores for selected object proposals and do it in a biological real-time - shorter than visual fixation time. 

In the following section we present our general framework for saliency driven object recognition with Deep CNN. 

\section{General approach for saliency driven object recognition with Deep CNN}\label{Sec:3}

Our method is developed for humans grasping an object. The task consists in simultaneous recognition of the object the subject wishes to grasp.  A human subject is instrumented with a glass-worn eye-tracker with a scene camera (Tobii Pro glasses 2). The block diagram of the method is presented in figure \ref{Fig1}. The data preparation block serves to filter out missing eye-tracker recordings, see Section \ref{Sec:32}. On the basis of recorded gaze fixations we compute visual saliency of pixels in the video recorded by the scene camera, see Section \ref{Sec:33}. In real world scenario our system fulfills automatic patch selection, i.e. "object proposal" (see the lowest branch of figure \ref{Fig1}) using computed saliency. Then CNN classifies the extracted patch into a set of known categories. Fusion of classification scores along the time allows for filtering natural noise such as eye-blinking and saccadic motion towards distractors. The middle branch of the diagram in figure \ref{Fig1} presents training process. To train the known category of objects we use a semi-automatic annotation method, also guided by computed saliency, see Section \ref{Sec:52}. Specific data augmentation approach is designed with regards to the real life scenario, see section \ref{Sec:422}. CNN training is realized on the augmented object dataset. In the following part of this section we detail these steps.

\tikzstyle{rond1} = [rectangle, rounded corners, minimum width=3cm, minimum height=0.6cm,text centered, draw=black]
\tikzstyle{rond2} = [rectangle, rounded corners, minimum width=3cm, minimum height=0.6cm,text centered, draw=blue, text=blue]
\tikzstyle{rond3} = [rectangle, rounded corners, minimum width=3cm, minimum height=0.6cm,text centered, draw=red, text=red]
\tikzstyle{rect2} = [rectangle, minimum width=3cm, minimum height=1cm, text centered, draw=black]
\tikzstyle{rect3} = [rectangle, minimum width=3cm, minimum height=1cm, text centered, draw=blue, text=blue]
\tikzstyle{rect4} = [rectangle, minimum width=3cm, minimum height=1cm, text centered, draw=red, text=red]

\tikzstyle{arrow1} = [thick,->,>=stealth]
\tikzstyle{arrow2} = [thick,->,>=stealth, draw=blue, text=red]
\tikzstyle{arrow3} = [thick,->,>=stealth, draw=red, text=blue]

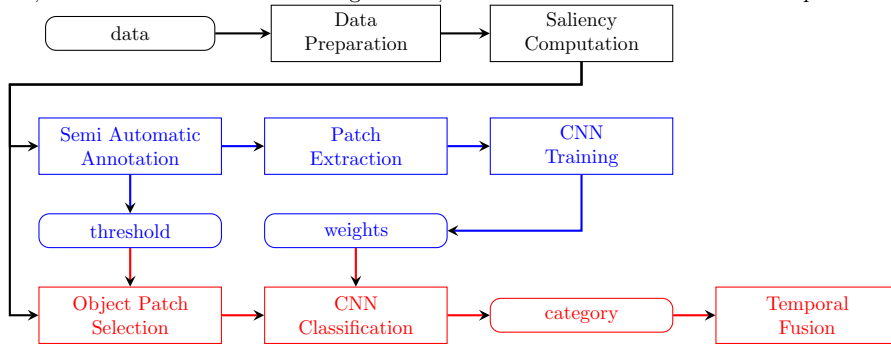
\begin{figure}
\caption{Block diagram of our method, the upper-branch blocks are common for training and test, the midle branch is the training scheme, and the lowest branch is the  online processing.}
\label{Fig1}
\centering
\begin{tikzpicture}[node distance=2cm]
	
	\node (data)     [rond1, scale=0.75] {data};
    \node (DataPrep) [rect2, scale=0.75, right of=data,     xshift=2.0cm, text width=2cm] {Data Preparation};
    \node (SaliComp) [rect2, scale=0.75, right of=DataPrep, xshift=2.0cm, text width=3cm] {Saliency\\ Computation};
    
    \node (SemiAnno) [rect3, scale=0.75, below of=data,     yshift=0.0cm, text width=3cm] {Semi Automatic\\ Annotation};
    \node (PatchExt) [rect3, scale=0.75, right of=SemiAnno, xshift=2.0cm, text width=3cm] {Patch\\ Extraction};
    \node (CNNTrain) [rect3, scale=0.75, right of=PatchExt, xshift=2.0cm, text width=3cm] {CNN\\ Training};
    
    \node (threshol) [rond2, scale=0.75, below of=SemiAnno, yshift=0.5cm, text width=3cm] {threshold};
    \node (weights)  [rond2, scale=0.75, right of=threshol, xshift=2.0cm, text width=3cm] {weights};
    
    \node (AutoAnno) [rect4, scale=0.75, below of=threshol, yshift=0.5cm, text width=3cm] {Object Patch\\ Selection};
	\node (CNNClass) [rect4, scale=0.75, right of=AutoAnno, xshift=2.0cm, text width=3cm] {CNN\\ Classification};
    \node (category) [rond3, scale=0.75, right of=CNNClass, xshift=2.0cm, text width=3cm] {category};
	\node (TmpFusio) [rect4, scale=0.75, right of=category, xshift=2.0cm, text width=3cm] {Temporal\\ Fusion};
    
    \draw [arrow1] (data) -- (DataPrep);
    \draw [arrow1] (DataPrep) -- (SaliComp);
    \draw [arrow1] (SaliComp.south) |- ++(0,-0.3) |- ++(-7.6,0) |- (SemiAnno.west);
    \draw [arrow1] (SaliComp.south) |- ++(0,-0.3) |- ++(-7.6,0) |- (AutoAnno.west);
    
    \draw [arrow2] (SemiAnno) -- (PatchExt);
    \draw [arrow2] (PatchExt) -- (CNNTrain);
    
    \draw [arrow2] (SemiAnno) -- (threshol);
    \draw [arrow2] (CNNTrain) |- (weights);
    
    \draw [arrow3] (AutoAnno) -- (CNNClass);
    \draw [arrow3] (CNNClass) -- (category);
    \draw [arrow3] (category) -- (TmpFusio);
    
    \draw [arrow3] (threshol) -- (AutoAnno);
    \draw [arrow3] (weights)  -- (CNNClass);
\end{tikzpicture}
\end{figure}

\subsection{Physiology of visual attention}\label{Sec:31}
In order to provide rationale for our data preparation methodology we briefly expose here our physiological hypotheses and the known neuro-physiological models of human vision relevant to our problem. 
Our actor, an upper-limb amputee has less freedom in the control of his body than a healthy person. When he wants to grasp an object in his environment he first looks at it (which is not always the case for a healthy subject). This is our main assumption.

The observation of a scene comprises: \textbf{(1)} The \textit{discovery} of the scene where the eye scouts sparsely the scene. \textbf{(2)} The \textit{fixation} on the object of interest. \textbf{(3)} \textit{Micro-saccades}, when the eye slightly oscillates about the target object. \textbf{(4)} \textit{grasping movement} is triggered. \textbf{(5)} Also some \textit{distractors}, such as audio; light; motion; and occlusions in the scene can lead the eye to deviate to another object.

We have conducted psycho-visual experiments on healthy volunteers aged from 20 to 23 and we observed that: \textbf{(1)} The \textit{scene discovery} takes from 240 to 300 ms. \textbf{(2)} The \textit{fixation} is about 250 ms. \textbf{(3)} \textit{Micro-saccades} can occur with  duration about 6 to 300 ms according to different sources reviewed in \cite{Martinez2009}, note that the frequency of our eye-tracker does not allow precise measurments of micro-saccades duration, see section \ref{Sec:511} for experimental set-up. \textbf{(4)} The \textit{grasping movement} then takes between 400 to 900ms. \textbf{(5)} Finally the times of \textit{distractor fixations} are between  100 and 500 ms. These data are in accordance with the results in \cite{Sal:Guerin}, \cite{Sal:Art}.

During the  \textit{scene discovery} a subject explores the scene searching for the target object, hence the object-of-interest is not fixated. Therefore we reject the beginning of each video sequence when selecting both training and validation frames and do this at the test stage as well. \textit{Micro-saccades} are not a problem thanks to the interpolation of fixation coordinates that maintain the eye fixation on the object-of-interest. \textit{Distractors}, nevertheless, are a real challenge: they cannot be automatically identified neither when performing semi-automatic annotation for training nor at the test stage.  And so they are included in our training set in the form of incorrectly labeled patches. To deal with them in our online framework, we propose to use temporal fusion of the classification results that filters out frames with distractors. 

\subsection{Data Preparation}\label{Sec:32}

The data preparation block receives the data from the Tobii Glasses 2 streams, simulated in the present work as two pre-recorded files: one for the video and one for the eye-tracking data. In the eye-tracking data some recordings of gaze fixations are missing due to eye-blinking of the subject. Also the video and the gaze tracking have different sampling rates, and so it is rare that an eye-tracking record is synchronized with a video frame.

To cope with all these problems, we apply interpolation (using a spline). This smoothes the gaze fixation data and allows us to synchronize the two streams.

\subsection{Subjective saliency Computation}\label{Sec:33}

\begin{figure}
    \caption{Different forms of Wooding's map (in raster scan order) \textbf{(i)} The original image $I$ with the fixation point $f$ drawn in red on it and
    \textbf{(ii)} Normalized Wooding's Map $W(I,f)$
    \textbf{(iii)} The Heat-Map visualization;
    \textbf{(iv)} The Weighted Map where the saliency is used to weight brightness in the frame.}
    \label{Fig2}
	\includegraphics[width=\linewidth]{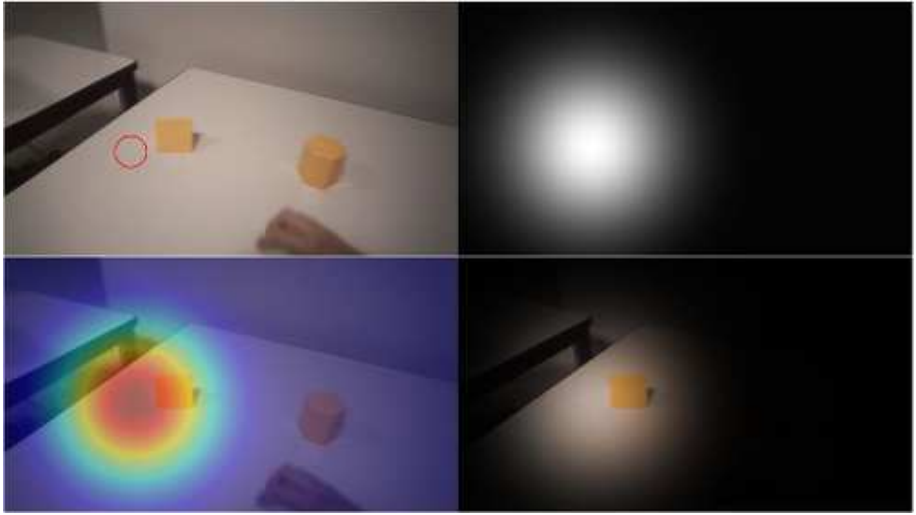}
\end{figure}

From the eye-tacker data we get  the recorded gaze-fixation point $f$ for the image $I$, with coordinates $x_f$, $y_f$ and $d_f$. Here $d_f$ is the coordinate along the axis of gaze direction and x, y are the coordinates of the fixation point in the image plane of video recorded with the scene camera of glasses. Therefore,  the so-called "subjective saliency map" or Wooding's map  $W(I,f)$ \cite{Sal:Wooding} can be computed. It is a normalized Gaussian function, centered on a fixation point, with values close to 1 in the vicinity of the fixation point (in the focal vision), and values close to 0 in pixels situated  far from it (in the peripheral vision). The spread  $\sigma(d)$ of the Gaussian is adapted to the size of the image $I$ and the distance $d_f$ to the object to model the focal vision. It is also normalized to sum to 1. A visual example is given in figure \ref{Fig2}. The equations \ref{Eq:WoodingMap} below detail the computation of the Wooding Map, and its parameters:
\begin{align}
	\sigma(I,d) &= \frac{A}{d} \cdot \frac{width(I) tan(180\alpha\pi)}{2 tan( \beta\pi / 180 )}\\
	W(I, f, x, y) &= \frac{A}{\norm{W}_\infty} \times \exp{\frac{-(x-x_f)^2 - (y-y_f)^2}{2\cdot \sigma(I,d_f)^2+\epsilon}}
	\label{Eq:WoodingMap}
\end{align}
Where $\alpha = 2^\circ$ is the angle of projection of the fovea, $\beta = 24^\circ$ is the camera opening angle on the width, $A = 1600$ mm is the maximum distance to an object according to our setting and $\epsilon$ is a small number $\epsilon=0.01$.

In our case of moving eye-tracker wearer, unlike the traditional Wooding's map, the spread $\sigma$ of the Gaussian function is linearly adapted wrt the distance to the object. Thus our method produces a  larger image patch when the object is closer (and so appear larger in the video), or reciprocally, smaller image patch when the object is farther. In figure \ref{Fig2} we show different forms of Wooding's saliency map on a video frame. 


\section{Network design and tuning}\label{Sec:4}

In our work we used the basic ImageNet architecture proposed in \cite{CNN:ImageNet}. We do not need a specific optimization of computational cost as the focused selection of an "object proposal" accordingly to the terminology of Girshik \cite{CNN:Girshick15}, in each frame allows us to be compatible with real-time requirements for object recognition. 
In this section we remind the architecture and focus on the way we extract "object proposal" patches and the background candidates based on saliency, and then augment them to prevent over-fitting.

\subsection{General architecture and parameters}\label{Sec:41}

\paragraph{Network layers}

The ImageNet network \cite{CNN:ImageNet}\ref{Tab1} is mainly composed of 5 types of layers: 5 convolutions (Conv), 3 fully connected (FC), 7 Rectified Linear Units (ReLU), 3 Max pooling, and 2 Local Response Normalization. They are combined vertically to increase the network depth.
\begin{enumerate}
\item{The convolution is used to extract features on its input by applying filter to it. On the first layer this filters respond to edges or color blobs, while on the last one they are able to abstract shapes and objects parts \cite{CNN:ImageNet}.}
\item{The FC layers are used to progressively map activation maps to a single dimension feature vector where at the end each value is associated to a class of object. It is then normalized to a probability distribution using a Soft Max layer.}
\item{The max polling \cite{CNN:ImageNet} is used to spatially down-sample the activation of the previous layer by only propagating the maximum activation of a previous group of locally connected neurons.}
\item{Local Response Normalization is used to normalize the response of neurons at the same spatial location. This is inspired by lateral inhibition in real neuron \cite{CNN:ImageNet}.}
\item{ReLU were introduced in \cite{CNN:ImageNet} to increase the network optimization} convergence.
\end{enumerate}

\begin{table}
	\caption{ImageNet \cite{CNN:ImageNet} architecture. $k$ is the kernel size, $nb$ is the number of filters learned, $b$ is the bias, $pad$ is the zero-padding size, and $s$ the stride}
	\label{Tab1}
    \tiny
\begin{center}
\begin{tabular}{|c|c|c|c|l|c|}
    \hline
    \textbf{Layer} & \textbf{Depth} & \textbf{Type} & \textbf{Name} & \textbf{Parameters} & \textbf{Top shape} \\ \hline
23 & 8 & Soft Max    & prob  &                                         & C         \\ \hline
22 & 8 & FC          & ip8   &                                         & C         \\ \hline
21 & 7 & Dropout     & drop7 & $ratio=0,5$                             & 4096      \\ \hline
20 & 7 & ReLU        & relu7 &                                         & 4096      \\ \hline
19 & 7 & FC          & ip7   & $b=1$                                   & 4096      \\ \hline
18 & 6 & Dropout     & drop6 & $ratio=0,5$                             & 4096      \\ \hline
17 & 6 & ReLU        & relu6 &                                         & 4096      \\ \hline
16 & 6 & FC          & ip6   & $b=1$                                   & 4096      \\ \hline
15 & 5 & Max pooling & pool5 & $k=3$x$3,~ s=2$                         & 6x6x256   \\ \hline
14 & 5 & ReLU        & relu5 &                                         & 13x13x256 \\ \hline
13 & 5 & Convolution & conv5 & $k=3$x$3,~ nb=256,~ pad=1,~ b=1$        & 13x13x256 \\ \hline
12 & 4 & ReLU        & relu4 &                                         & 13x13x384 \\ \hline
11 & 4 & Convolution & conv4 & $k=3$x$3,~ nb=384,~ pad=1,~ b=1$        & 13x13x384 \\ \hline
10 & 3 & ReLU        & relu3 &                                         & 13x13x384 \\ \hline
9  & 3 & Convolution & conv3 & $k=3$x$3,~ nb=384,~ pad=1,~ b=0$        & 13x13x384 \\ \hline
8  & 2 & LRN         & norm2 & $k=5$x$5,~ \alpha=10^{-4},~ \beta=0,75$ & 13x13x256 \\ \hline
7  & 2 & Max pooling & pool2 & $k=3$x$3,~ s=2$                         & 13x13x256 \\ \hline
6  & 2 & ReLU        & relu2 &                                         & 27x27x256 \\ \hline
5  & 2 & Convolution & conv2 & $k=5$x$5,~ nb=256,~ pad=2,~ b=1$        & 27x27x256 \\ \hline
4  & 1 & LRN         & norm1 & $k=5$x$5,~ \alpha=10^{-4},~ \beta=0,75$ & 27x27x96  \\ \hline
3  & 1 & Max pooling & pool1 & $k=3$x$3,~ s=2 $                        & 27x27x96  \\ \hline
2  & 1 & ReLU        & relu1 &                                         & 55x55x96  \\ \hline
1  & 1 & Convolution & conv1 & $k=11$x$11,~ nb=96,~ s=4,~ b=0$         & 55x55x96  \\ \hline
0  & 0 & Data        & data  &                                         & 227x227x3 \\ \hline
\end{tabular}
\end{center}
\end{table}

\paragraph{Optimization}

We use the soft max loss function (multinomial logistic loss) already implemented in Caffe \cite{Web:Caffe}, that for input image $X_i$ with known label $l_i$ is:\\
\begin{align}
E(X_i,~l_i) &= -\frac{1}{N} \sum_{j=1}^{N} log(\widehat{P}_j)\delta(\widehat{l}_j,l_i) \\
\delta(l,l_i) &= {~ 1 \text{ if } l = l_i,~ 0 \text{ else} }
\end{align}
Where $\widehat{P}_j$ is the probability with associated label $\widehat{l}_j$, resulting from the forward pass in the network. The overall loss over a dataset $D$ is:
\begin{equation}
L(W) = \frac{1}{\card{D}} \sum_{i=1}^{\card{D}} E(X_i, l_i) + \lambda r(W)
\end{equation}
\small$r(W)$ is a regularization term with weight decay $\lambda=0.0005$

We use the default implementation of stochastic gradient descent from Caffe and ImageNet \cite{CNN:ImageNet,Web:Caffe}  with weight update rule:
\begin{align}
	V_{t+1} &= \mu V_t - \alpha \nabla L(W_t) \\
    W_{t+1} &= W_t + V_{t+1}
\end{align}
Where $\alpha$ is the Learning rate and $\mu$ the Momentum (0.9). Our learning rate $\alpha$ is initialized to $0.001$ and decreased by half every $30000$ iterations.
$\nabla L(W_t)$ is computed through back-propagation in the network. 

\subsection{Saliency-based data preparation}\label{Sec:42}

We will now present the preparation of the input data for the network training, which consists in selection of bounding boxes of objects using saliency maps, and sampling of background patches. We also propose a data augmentation strategy corresponding to our problem.

\subsubsection{Patch extraction}\label{Sec:421}

Our basic CNN \cite{CNN:ImageNet} has a fixed input resolution, so no matter at which resolution the objects appear in the video, they are all resized to a fixed size (227x227 RGB in our case). For machine learning, especially classification, we have to extract object patch (examples) for all categories, including the background that is the rejection category. It is important to extract a similar amount of examples for each of them to avoid imbalanced class problem.

\begin{figure}
    \caption{Patch extraction: \textbf{(i)} Middle: bounding box of the object-of-interest \textbf{(ii)} Middle left and right: the exclusion area. \textbf{(iii)} Top and bottom: example of background patches with a maximum overlap of  20\%}
    \label{Fig3}
    \includegraphics[width=\linewidth]{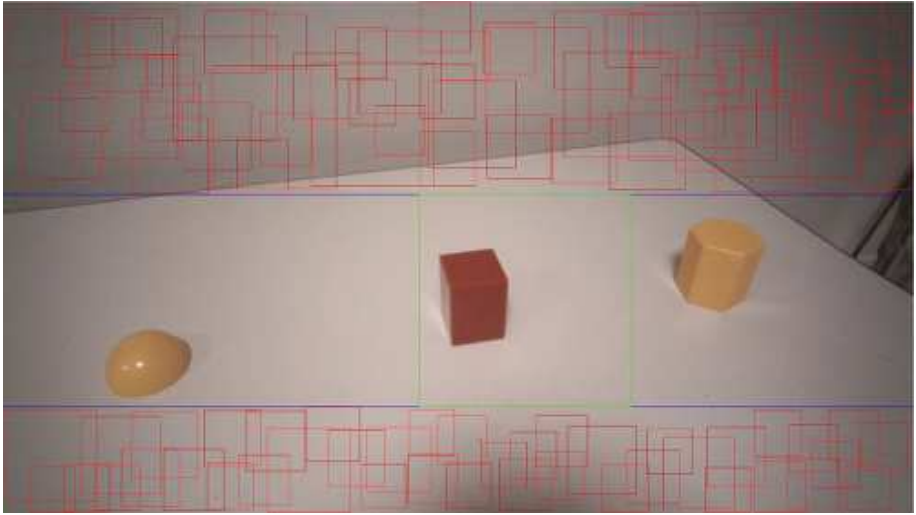}
\end{figure}

In our scenario, the object of attention is identified by the thresholded saliency, and a label describing the category. We extract the corresponding blob by connected  component analysis, it gives us the bounding box.
The green bounding box in figure \ref{Fig3} is an example of an image patch corresponding to a "rectangular prism". 
We also have to extract background at the same time to ensure that we have the same amount of rejection examples. Remember that our experimental setup specifies that the objects are lined-up on the table. Since only one object is labeled, we exclude the bounding box of the object, but also the area where other objects could appear, to avoid a background/object mixture. This exclusion area is drawn in blue in the figure  \ref{Fig3} below.

Background patches are then sampled randomly in the remaining parts of the image. Their minimum resolution is limited to 95x95 pixels induced by full HD resolution of our videos (1080x1920). When sampling several background patches in the same image, we respect the maximal overlap of 20\%. Thus we ensure sampling of  different areas in the image background and therefore, we capture more information on it. The figure \ref{Fig3} shows many random background patch proposals. In practice we  keep only one or two per video frame in order to avoid an imbalanced class problem.  Due to the random sampling and the repeatability of the background in video, we extract samples well covering the background of video scenes. 

\subsubsection{Patch Augmentation}\label{Sec:422}

\begin{figure}
	\caption{Patch augmentation: Raws show different object categories and column show different augmentations. Each group of 4 columns depict different rotation angle $\alpha$, and within this, each column is a different blurring kernel size $k$}
    \label{Fig4}
	\includegraphics[width=\linewidth]{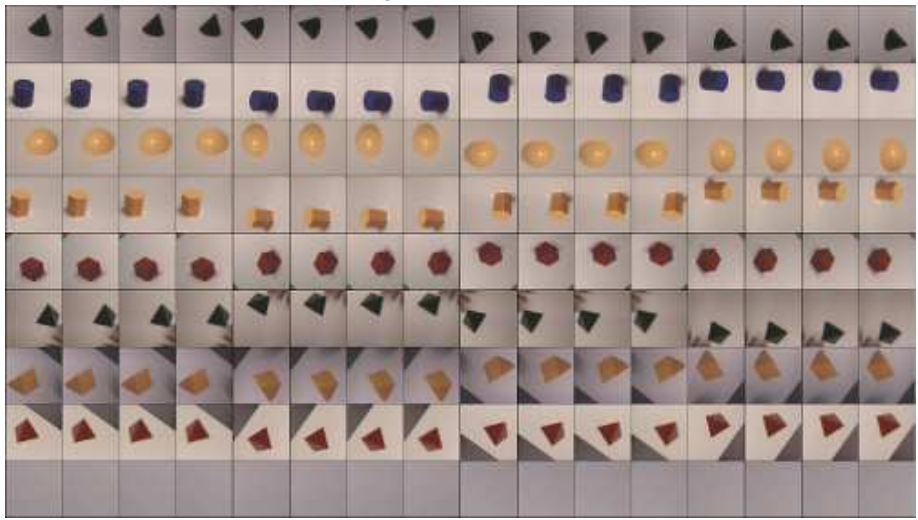}
\end{figure}

Data augmentation is very efficient to prevent over-fitting \cite{CNN:ImageNet}. The idea is to apply label-preserving transformation to the image patch, and give both the original one and the transformed ones to the network for training. This will artificially increase the training dataset size. Common transformations are horizontal mirroring, random cropping \cite{Web:Caffe}. In our case objects can be placed upside-down or lay on any of their sides, this led us to rotate training image patch by an angle $\alpha \in \{0^\circ, 90^\circ, 180^\circ, 270^\circ\}$. We only considered multiples of $90^\circ$ to avoid discretization problems that can lead to a drastic accuracy drop, due to the parasite high-frequency components in the image spectrum. As the video can be blurred by fast motion of the glass-worn camera, which is often the case in egocentric videos, we decided to blur training image patches by 3 Gaussian kernels of size $k \in \{1\times1, 3\times3, 5\times5, 7\times7\}$. This increases the network robustness to motion blur. In total we increase our training by $16$ $((4\times N)\times 4)$ (the rotation by $\alpha=0^\circ$ followed by a blurring $k=1\times1$ leave a patch unchanged). We also apply this transformation to background image patches to preserve class balance. The figure \ref{Fig4} shows an example for all object categories.

Note, that we do not need data augmentation at the test step as it is proposed in ImageNet \cite{CNN:ImageNet}. They need it as they do not have certainty on the object location. This is why they generate multiple candidates for the "object proposal" and practice fusion of scores. In our case an "object proposal" is unique as it is totally defined by online recorded gaze fixation and derived saliency maps.

\subsection{Temporal Fusion}\label{Sec:423}

We are classifying a sequence of object proposals in a video  with "mean fusion" operator, which is equivalent to a simple sum as shown in the equation below \ref{Eq:TempFusion}: The scores for each class are summed over the candidate patches along a video, and the final category is the one that obtains the maximum score over these sums.
\begin{equation}
	c(V) = \argmax_{c}{s(V,~ c)} =~ \argmax_{c}\{\sum_{i=1}^{N}{s(p_i,~ c)}\}
    \label{Eq:TempFusion}
\end{equation}
Where $V$ is a video, $p_i$ is the candidate patch of frame $i$. 

We use this instead of retaining just the most frequent classification result, i.e. the "majority vote", because we believe the score of an incorrectly classified patch is often much smaller than the score of a correctly classified patch. And so by summing the scores over a patch, the correct class score wins.

\section{Experiments and results}\label{Sec:5}

In order to test our object recognition framework in a real life but simplified scenario we produced a new dataset that we called Large Egocentric Gaze Objects (LEGO) which will be soon available online. Below we present the experimental setup and content; our data selection results, network optimization, and results. 

\subsection{LEGO Dataset}\label{Sec:51}

\subsubsection{Experimental Setup}\label{Sec:511}

\begin{figure}
	\caption{Left: A subject equipped with the Tobii Pro Glasses 2 performing the experiment. Right: The egocentric field of view of the glasses.}
	\label{Fig5}
    \centering
	\includegraphics[height=3.4cm]{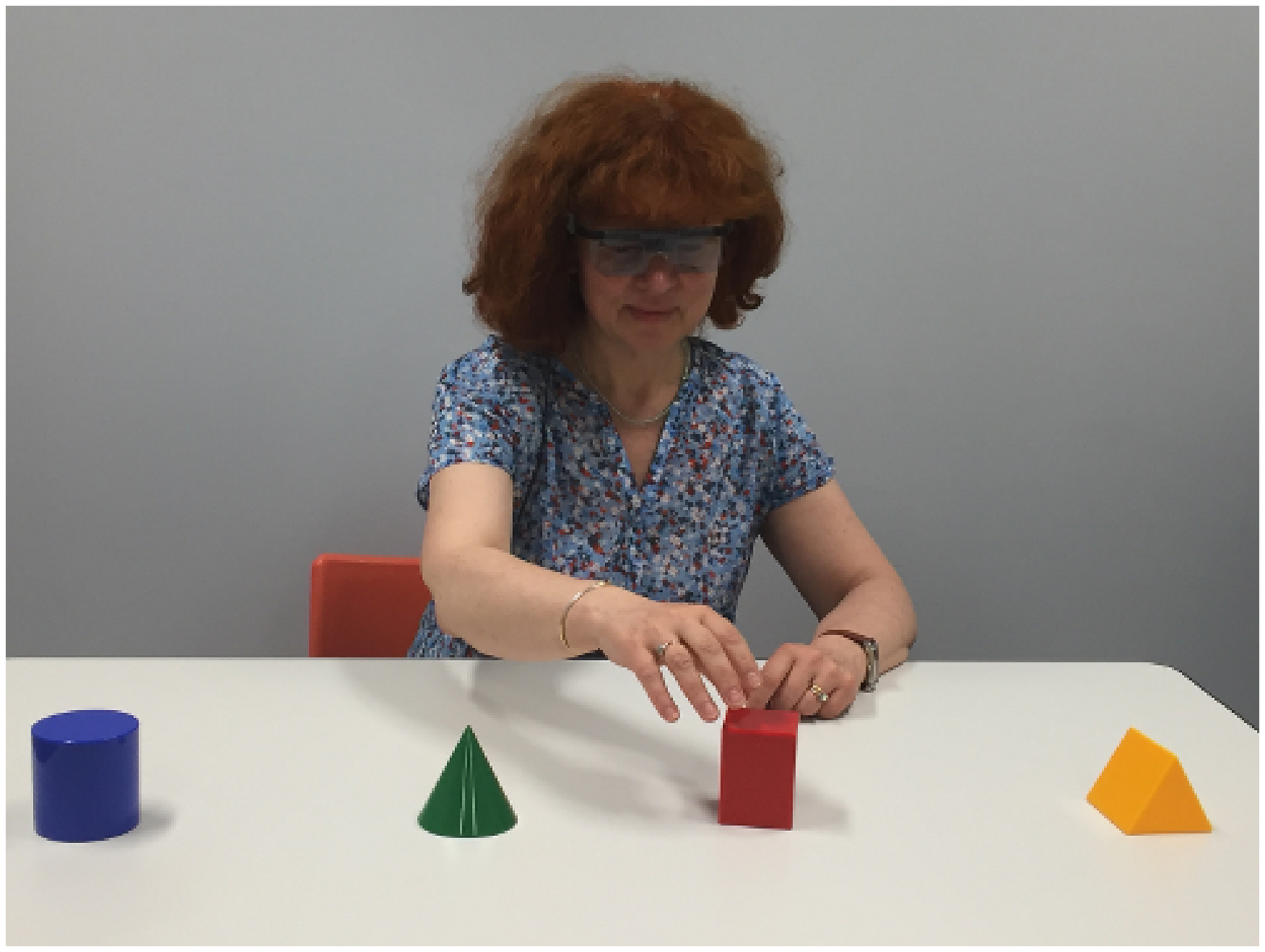}\hspace*{0.1cm}
    \includegraphics[height=3.4cm]{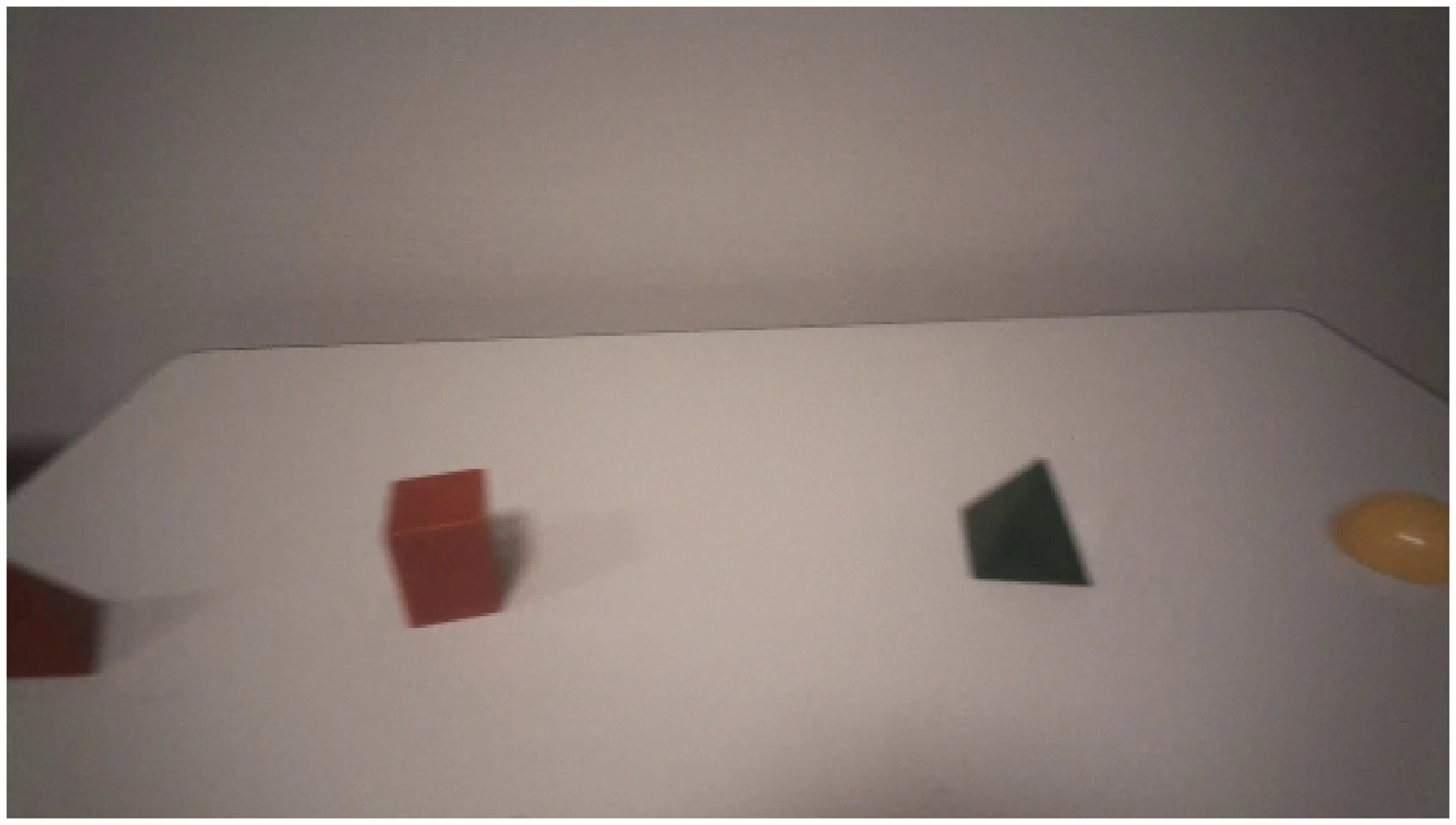}
\end{figure}

The recording of our dataset was conducted with four healthy volunteers aged from 20 to 23. In each recording session a subject was instructed to look for a specific object and to grasp it. The subjects were sitting in front of a white table, facing a white wall. They wore the Tobii Pro Glasses 2. This eye-tracker records gaze fixations at 50 Hz and video frames at 25 Hz. At first, the subject's eyes were closed (in which case the gaze data are not available). Four objects out of eight different objects were randomly chosen and placed in line on the table. They were presented in different positions for each experiment (they could be placed upside down, flipped, and rotated). The name of the object to grasp was revealed at this moment and the video recording started at the same time. The subject could then open his eyes, search for the object, and once he found it he grasped it. After a few seconds the recording was stopped. Figure \ref{Fig5} shows a subject performing the experiment.

\subsubsection{Videos}\label{Sec:512}

We recorded 151 videos with our experimental setup \ref{Sec:511}. Eight types of objects with simple shape and identifiable color were used, see examples in figure \ref{Fig5}. The duration of videos was between 3,6 s up to 11,9 s, that is 6,5 $\pm$ 0,9 s on average. These videos are short as they depict the initiation of hand motion and grasping of the object-of-interest. They are split between Train ($60\%$), Validation ($20\%$) and Test ($20\%$) sets as shown in table \ref{Tab2}.

This video dataset is rather simple in a sense that the object and the background are well separable; there are no occlusions. Some motion blur is observed when the subject moved his head when searching for the object-of-interest. The real challenge of this dataset comes from the semi-automatic ground truth annotation and distractors. Indeed some frames are miss-annotated: the object in the video can have an incorrect label as the subject was distracted and did not fixate the right object. The localization can be inaccurate; the bounding box can be too big or too small due to distance measurement inaccuracy in saliency map computation.

\begin{table}[!ht]
	\caption{Number of videos in the LEGO dataset, by category for Train, Validation and Test sets.}
	\label{Tab2}
	\centering
\begin{tabular}{ccrrrr}
	\hline
    \multicolumn{2}{c}{\textbf{Categories}} & \textbf{Training} & \textbf{Validation} & \textbf{Testing} & \textbf{Total} \\ \hline
	~ & Background & 90 & 31 & 0 & 121  \\
	\includegraphics[width=0.3cm]{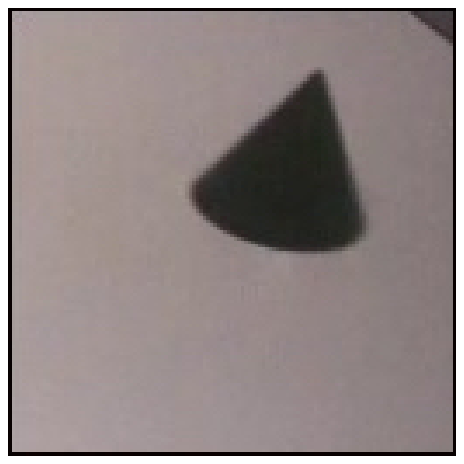} & Cone & 13 & 5 & 4  & 22 \\
	\includegraphics[width=0.3cm]{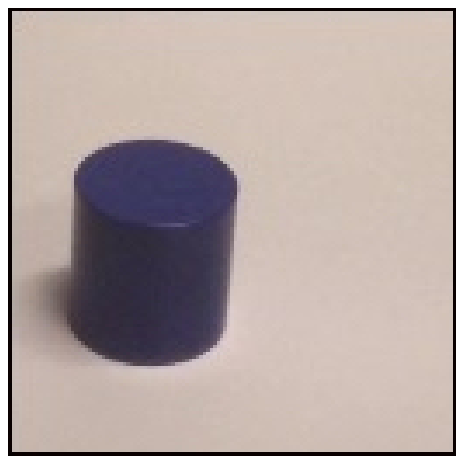} & Cylinder & 4 & 2 & 1 & 7 \\
	\includegraphics[width=0.3cm]{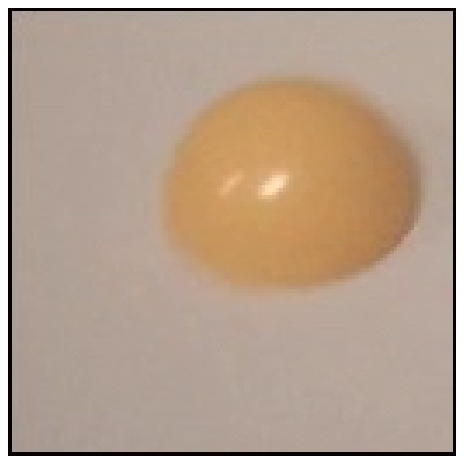} & Hemisphere & 8 & 3 & 3 & 14 \\
	\includegraphics[width=0.3cm]{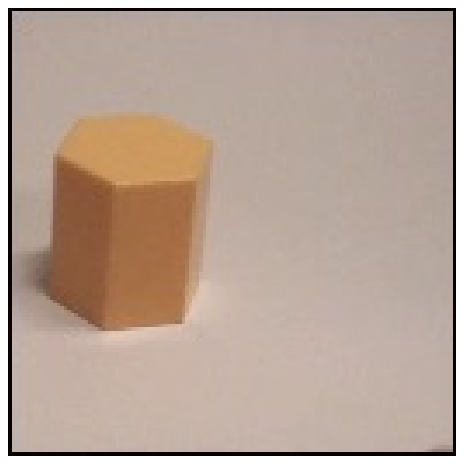} & Hexagonal\_Prism & 10 & 4 & 3 & 17 \\
	\includegraphics[width=0.3cm]{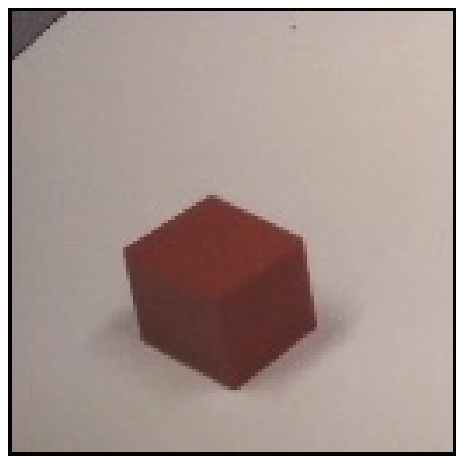} & Rectangular\_Prism & 17 & 5 & 6 & 28 \\
	\includegraphics[width=0.3cm]{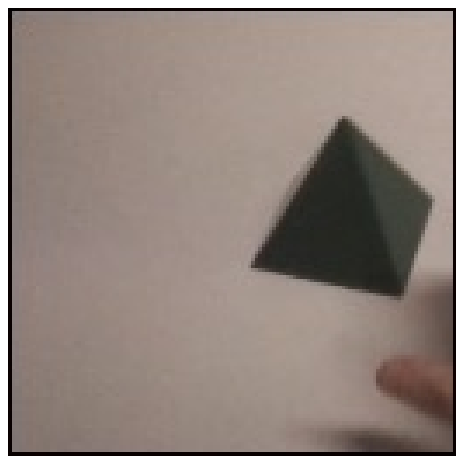} & Rectangular\_Pyramid & 10 & 4 & 3 & 17 \\
	\includegraphics[width=0.3cm]{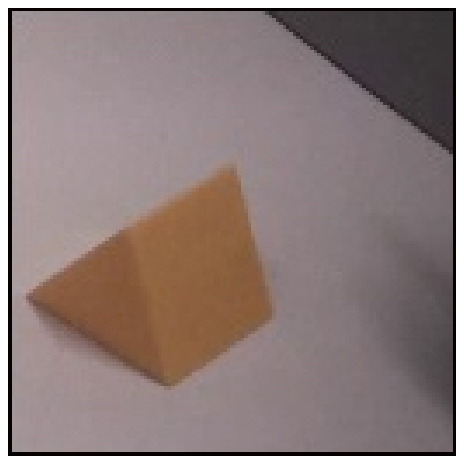} & Triangular\_Prism & 17 & 5 & 4 & 26 \\
	\includegraphics[width=0.3cm]{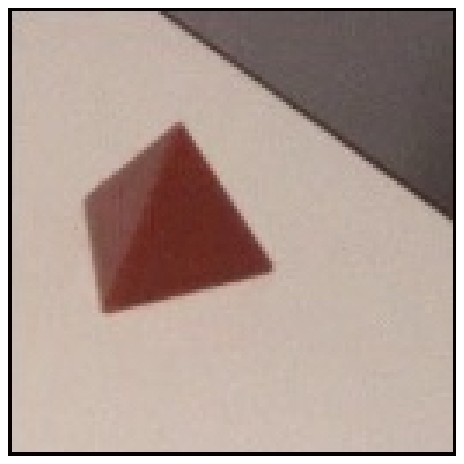} & Triangular\_Pyramid & 11 & 3 & 4 & 18 \\ \hline
    \multicolumn{2}{c}{\textbf{Total}/BGD} & \textbf{90} & \textbf{31} &   \textbf{28} & \textbf{149} \\ \hline
\end{tabular}
\end{table}


\subsection{Semi Automatic Annotation}\label{Sec:52}

Annotation is the process of describing the content of a set of videos, in each frame of each video, for all type of content (objects). In computer vision, this is done manually: a human annotator visualizes the videos, select objects in each frame depicting their bounding boxes (rectangle+label) and sometimes segments the object (binary mask+label). Datasets are now on the order of millions of images, and thousand of object types turning annotation into enormous and tedious work.

In our experiment we know the subject looked at the object, so we propose to use the saliency map to select the patch of the object (the saliency peak is located on the object). We threshold the saliency to create an approximate segmentation mask. In practice, their is a delay between the beginning of the video and the moment the subject opens his eyes, and finds the object of interest, called visual exploration (see section \ref{Sec:31}. It is on the order of 300 ms. We have to ignore this part of the sequence or we would be considering patches of objects other than the object of attention.

\begin{figure}
	\caption{Annotation Tool user interface: the sequence can be played using buttons 2 to 4; the visualization method and resolution can be selected with buttons 5 and 6; the annotation parameters with buttons 7 to 11.}
    \label{Fig6}
	\includegraphics[width=\linewidth]{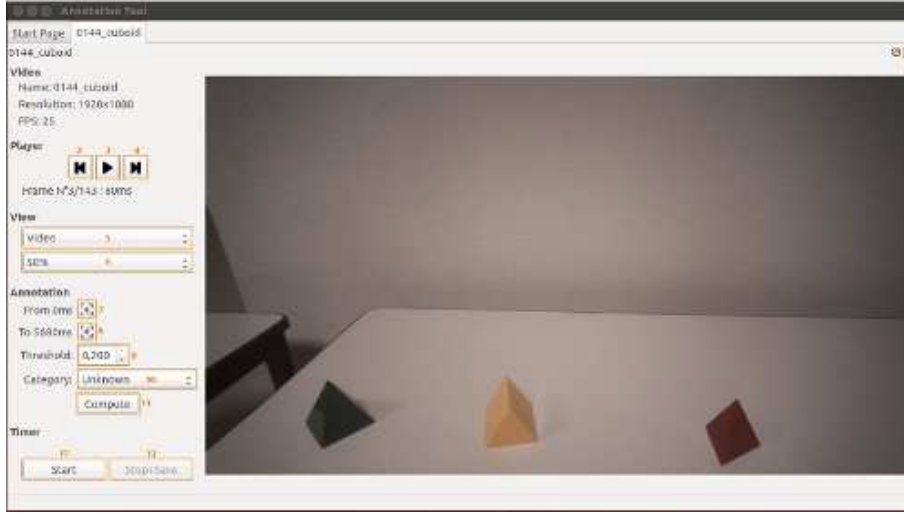} 
\end{figure}

To solve this problem we developed a simplified annotation tool presented in figure \ref{Fig6}, that allows the human annotator \textbf{(1)} to select the moment when the scene exploration is completed and the subject is focused on the object-of-interest (buttons 7 and 8), \textbf{(2)} to threshold the saliency map (button 9), and \textbf{(3)} to choose the category of the object (button 10). We allow the threshold to be changed because in some sequences the saliency map is larger due to imprecision in the distance to the object-of-interest computed by the eye-tracker software. Changing the thresholds allows the annotator to control the amount of context inside the object patch.

When the subject is distracted by other objects or lightning changes in the background the human annotator is not always able to detect this moment. This leads to the noise in training data and selection of a "bad" object proposal in online test scenario. This results in an accuracy drop we observed in our experiments. We deal with it using score fusion (see section \ref{Sec:423})

\subsection{Patch Extraction}\label{Sec:53}

We extracted an image patch as described in section \ref{Sec:421}. We only took one background patch on frames showing an object in the training and validation dataset, so that the final number of background patch sums up almost to the sum of those in object categories. We do not sample background on the test set. Ultimately, we had as many patches per category as the ImageNet dataset \cite{DataSet:ImageNet}, but fewer categories were considered: $9$ instead of $1000$.

\begin{table}
	\caption{Number of image patches by category for Train, Validation and Test extracted from the LEGO dataset.}
	\label{Tab3}
	\centering
\begin{tabular}{ccrrrr}
	\hline
    \multicolumn{2}{c}{\textbf{Categories}} & \textbf{Training} & \textbf{Validation} & \textbf{Testing} & \textbf{Total}\\\hline
	~ & Background & 123 424 & 43 024 & 0 & 166 448 \\
	\includegraphics[width=0.3cm]{Cone.eps} & Cone & 17 824 & 6 352 & 420 & 24 596 \\
	\includegraphics[width=0.3cm]{Cylender.eps} & Cylinder & 6 544 & 2 928 & 111 & 9 583  \\
	\includegraphics[width=0.3cm]{Hemisphere.eps} & Hemisphere & 13 360 & 4 016 & 272 & 17 648  \\
	\includegraphics[width=0.3cm]{HexagonalPrism.eps} & Hexagonal\_Prism & 16 592 & 5 776 & 235 & 22 603 \\
	\includegraphics[width=0.3cm]{RectangularPrism.eps} & Rectangular\_Prism & 24 032 & 6 816 & 620 & 31 468 \\
	\includegraphics[width=0.3cm]{RectangularPyramid.eps} & Rectangular\_Pyramid & 10 736 & 4 448 & 308 & 15 492 \\
	\includegraphics[width=0.3cm]{TriangularPrism.eps} & Triangular\_Prism & 21 168 & 7 744 & 396 & 29 308 \\
	\includegraphics[width=0.3cm]{TriangularPyramid.eps} & Triangular\_Pyramid & 16 784  & 4 976 & 412 & 22 172 \\\hline
    \multicolumn{2}{c}{\textbf{Total}} & \textbf{250 464} & \textbf{86 080} & \textbf{2 774} & \textbf{339 318} \\\hline
	\end{tabular}
\end{table}

\subsection{Network Optimization}\label{Sec:54}

We used the ImageNet architecture \cite{CNN:ImageNet}, changing the number of outputs of the last FC layer to match our number of classes. The learning rate was set to $0,001$ and was decreased every $30000$ iterations by half.

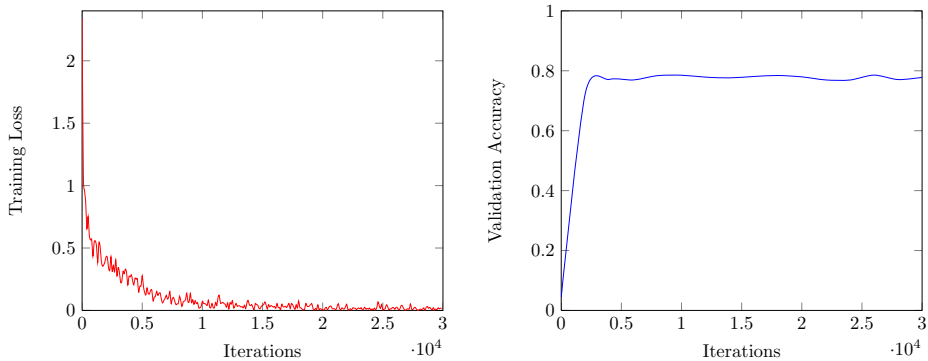
\begin{figure}
	\caption{Training and validation of the network. Left: Training loss and learning rate as a function of the number of iterations. Right: Validation accuracy as a function of the number of iterations}
    \label{Fig7}
    \centering
    \begin{tikzpicture}[scale=.7]
        \begin{axis}[
        	ymin=0, ymax=2.4,
        	xmin=0, xmax=30000,
            xlabel={Iterations},
            ylabel={Training Loss},
        ]
        \addplot[smooth,red] file { training.txt };
        \end{axis}
    \end{tikzpicture}\hspace*{.3cm}
    \begin{tikzpicture}[scale=.7]
        \begin{axis}[
        	ymin=0, ymax=1,
        	xmin=0, xmax=30000,
            xlabel={Iterations},
            ylabel={Validation Accuracy},
        ]
        \addplot[smooth,blue] file { validation.txt };
        \end{axis}
    \end{tikzpicture}
\end{figure}

The training loss is rapidly decreasing as shown in figure \ref{Fig7}:left The validation accuracy rapidly reaches $\sim 80\%$ as shown in figure \ref{Fig7}:right indicating a stable training of our network. This fast training is due to the simplicity of our objects and the lack of clutter in the scenes.

We train our network on a server equipped with 56 Intel Xeon cores and a NVIDIA Tesla k40m, it took two days to complete 47000 iterations.

\subsection{Classification results}\label{Sec:55}

\begin{figure}
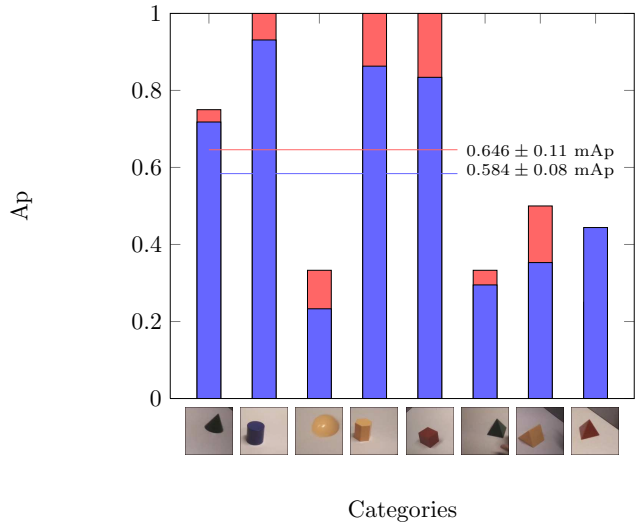

	\centering
	\begin{tikzpicture}[scale=0.9]
        \begin{axis}[
            xtick={1,2,3,4,5,6,7,8},
			xticklabels={
            	\includegraphics[width=0.7cm]{Cone.eps},
            	\includegraphics[width=0.7cm]{Cylender.eps},
            	\includegraphics[width=0.7cm]{Hemisphere.eps},
            	\includegraphics[width=0.7cm]{HexagonalPrism.eps},
            	\includegraphics[width=0.7cm]{RectangularPrism.eps},
            	\includegraphics[width=0.7cm]{RectangularPyramid.eps},
            	\includegraphics[width=0.7cm]{TriangularPrism.eps},
            	\includegraphics[width=0.7cm]{TriangularPyramid.eps}
            },
            ymin=0,
  			ymax=1,
            xlabel={Categories},
            ylabel={Ap},
            xlabel style={at={(axis description cs:0.5,-0.15)},anchor=north},
    		ylabel style={at={(axis description cs:-0.1,.5)},anchor=south}
        ]
        \addplot[ybar,fill=red!60] coordinates {
                (1, 0.750)
                (2, 1.000)
                (3, 0.333)
                (4, 1.000)
                (5, 1.000)
                (6, 0.333)
                (7, 0.500)
                (8, 0.250)
        };
        \addplot[ybar,fill=blue!60] coordinates {
                (1, 0.718)
                (2, 0.931)
                (3, 0.233)
                (4, 0.863)
                (5, 0.834)
                (6, 0.295)
                (7, 0.353)
                (8, 0.444)
        };
        \addplot[blue!60,samples=100,domain=1:5.5] {0.584};
        \node[anchor=south] at (axis cs:7,0.55) {\scriptsize $0.584\pm 0.08$ mAp};
        \addplot[red!60,samples=100,domain=1:5.5] {0.646};
        \node[anchor=south] at (axis cs:7,0.6) {\scriptsize $0.646\pm 0.11$ mAp};
        \end{axis}
	\end{tikzpicture}
    \caption{Average precision per class without (0,584 mAp), and with (0,646 mAP) score fusion}
    \label{Fig8}
\end{figure}
Figure \ref{Fig8} shows the average precision on all our categories as well as the gains of our temporal score fusion method. The deep CNN classifier was able to achieve the state-of-the-art performance with a mean average precision of $58,4\%$. Then score fusion presented in section \ref{Sec:423} was able to recover the correct category over a video sequence even if less than 40\% of the patches extracted on the frames were correctly classified. This was because the few correctly classified object proposals had strong score of their class. In the present work, we did fusion of scores on the whole video. For real time application in the assistive neuro-prosthesis visual system, the size of the fusion buffer would have to be optimized with respect to the latency of other components of the whole system. This leads to a score of $64,6\%$ mAp,yielding a $6,2\%$ gain.

We have conducted a performance test under Ubuntu 14.04 on an Intel I7-4790@3.6GHz CPU and a NVIDIA Quadro K4200 GPU. Computation of the saliency map from raw eye-tracking data was implemented in C++ with OpenCV and CUDA. This computation took 20 ms per frame without CUDA acceleration and 5 ms with it. We use the Caffe toolbox. It is not configured with CUDNN acceleration, so better performance can be expected with it. Classifying 2767 patches from 28 videos took 23.877 s which means 8.6 ms per patch, for a single video it means a computational time of 995 ms on average.

The total time for the computation of the saliency, and of classification of a patch, is of 28.6ms which is less than video frame rate (40ms) and much less than our requirement to be faster than a gaze-fixation time (250ms).

\section{Discussion, conclusion and perspectives}\label{Sec:6}

In this work, we have proposed an approach for object recognition in egocentric videos guided by visual saliency to help grasping actions for neuro-prostheses.  For annotation of visual data for traning of object detectors in such a setting we also proposed a semi-automatic annotation method, guided by visual saliency as well. The recorded egocentric dataset will soon be made available online. Object recognition 
is performed using a deep CNN \cite{CNN:ImageNet} that was able to achieve $58,4\%$ mAp, and using temporal fusion of scores, we obtained state-of-the art results of $64,6\%$ mAp despite the presence of annotation noise introduced by the distractors in training set and in the real-world "online" testing. The total time of our recognition system is about 28  ms per frame including visual saliency map computation and generalization with the Deep CNN.  This time  matches our requirement to be faster than visual fixation time.

This method has yet to be tuned for live system integration and some parameters such as the buffer size for temporal filtering have to be adjusted wrt the latency of other components of the neuro-prosthesis system. We are also eager to try other deeper and wider CNN architectures. The proposed approach gives rize to a wide set of exploration routes. Indeed, in such settings, the presence of noise in both annotation and test data sets are unavoidable due to human errors and physiology of human attention. We are thus interested in developing noise-robust optimization methods for Deep Neural Networks. 


\section*{Acknowledgments}

This work was supported by CNRS-Idex grant PEPS Suivipp 2015.
\section*{References}
\bibliography{Bibliography}

\end{document}